\documentclass[preprints,datadescriptor,accept,pdftex,moreauthors]{mdpi}

\usepackage[utf8]{inputenc}
\usepackage[T1]{fontenc}
\usepackage{lineno}
\usepackage{csvsimple}
\usepackage{graphicx}
\usepackage{subcaption}
\usepackage{mwe}

\firstpage{1}
\pdfoutput=1

\Title{MN-DS: A Multilabeled News Dataset for News Article
	 Hierarchical Classification}

\TitleCitation{MN-DS: A Multilabeled News Dataset for News Article Hierarchical Classification}

\Author{Alina Petukhova *\orcidA{}, Nuno Fachada \orcidB{}}

\AuthorNames{Alina Petukhova and Nuno Fachada}

\AuthorCitation{Petukhova, A.; Fachada, N.}

\address[1]{Lusófona University, COPELABS, Campo Grande 376, 1749-024
 Lisbon, Portugal; nuno.fachada@ulusofona.pt\\

}

\corres{\hangafter=1 \hangindent=1.05em \hspace{-0.82em}
Correspondence: alina.petukhova@ulusofona.pt}

\abstract{This article presents a dataset of 10,917 news articles with hierarchical news categories collected between  1 January 2019 and  31 December 2019. We manually labeled the articles based on a hierarchical taxonomy with 17 first-level and 109 second-level categories. This dataset can be used to train machine learning models for automatically classifying news articles by topic. This dataset can be helpful for researchers working on news structuring, classification, and predicting future events based on released news.}


\begin{document}

\section{Background and Summary}

A news dataset is a collection of news articles classified into different categories. In the past decade, there has been a sharp increase in news datasets available for analysis~\cite{nlpdatasetsoverview}. These datasets can be used to understand various topics, from politics to the economy.

A few different types of news datasets are commonly used for analysis. The first is raw data, which includes all the data that a news organization collects. This data can be used to understand how a news organization operates, what stories are covered, and how they are covered. The second type of news dataset is processed data. These data have been through some processing, such as aggregation or cleaned up. Processed data are often easier to work with than raw data and can be used to answer specific questions such as providing additional information for the decision-making process. The third type of news dataset is derived data. These data are created by combining multiple datasets, often from different sources~\cite{federatednewsclassification}. News datasets can be used for various purposes in a machine learning context, for example:
\begin{itemize}
\item Predicting future events based on past news articles.
\item Understanding the news cycle.
\item Determining the sentiment of news articles.
\item Extracting information from news articles (e.g., named entities, location, dates).
\item Classifying news articles into predefined categories.
\end{itemize}

To adequately answer research questions, news datasets should contain sufficient data points and span a significant enough period. There are many labeled news datasets available, each with specific limitations. For example, they may only cover a specific period or geographical area or be confined to a particular topic. Additionally, the categories may not be completely accurate, and the datasets may be biased in some way~\cite{Stefansson2014QuantitativeMO,gezici2022}.

Some of the more popular news datasets include the 20 Newsgroups dataset~\cite{newsgroups}, AG's news topic classification dataset~\cite{agnewsclassification}, L33-Yahoo News dataset~\cite{yahoodataset, yahoodataset2}, News Category dataset~\cite{newskaggledataset}, and Media Cloud dataset~\cite{mediaclouddataset}. Each of these datasets has been used extensively by researchers in the fields of natural language processing and machine learning, and each has its advantages and disadvantages. The 20 Newsgroups dataset was created in 1997 and contains 20 different categories of news, each with a training and test set. The data is already pre-processed and tokenized, which makes it very easy to use. However, the dataset is outdated and relatively small, with only about 1000 documents in each category.

The AG's news topic classification dataset is a collection of news articles from the academic news search engine ``ComeToMyHead'' during more than one year of activity. Articles were classified into 13 categories: business, entertainment, Europe, health, Italia, music feeds, sci/tech, software \& dev., sports, toons, top news, U.S., and world. The dataset contains more than 1 million news articles. However, there are several limitations to this dataset. First, it is currently outdated since data were collected in 2005. Second, the taxonomy covers specific countries such as the US and Italy but has general references such as Europe or world, creating overlaps in the classification (e.g., Italy and Europe) as well as potential imbalances (e.g., events in China are likely to be underrepresented and/or under-reported compared to those in the US). Finally, the dataset does not include methods for type or category description.

The L33-Yahoo News dataset is a collection of news articles from the Yahoo News website provided as part of the Yahoo! Webscope program. The articles are labeled into 414 categories such as music, movies, crime \and justice, and others. The dataset includes the random article id followed by possible associated categories. The L33-Yahoo News dataset is available under Yahoo's data protection standards. It can be used for non-commercial purposes if researchers credit the source and license new creations under identical terms. The limitations of the L33 dataset are the license terms, restricting companies from using this dataset for commercial purposes, and the amount of data per class, with the category ``supreme court decisions'' having only five articles, for example. In addition, there is some overlap in the categories, which makes it challenging to train a model that can accurately predict multiple categories.

The News Category Dataset is a collection of around 210k news articles from the Huffington Post, labeled with their respective categories, which include business, entertainment, politics, science and technology, and sports. However, the dataset has several limitations. First, the dataset is not comprehensive since it only includes articles from one source. Second, news categories are not standardized, including broad categories such as ``Media'' and ``Politics'' and very narrow ones like ``Weddings'' and ``Latino voices''.

The Media Cloud Data Set is a collection of over 1.7 billion articles from more than 60 thousand media sources around the world. The dataset includes articles from both mainstream and alternative news sources, including newspapers, magazines, blogs, and online news outlets. Data can be queried by keyword, tag, category, sentiment, and location. This dataset is useful for researchers who are interested in studying media coverage of specific topics or trends over time. Media Cloud is a large multilingual dataset that has good media coverage but limited use in topic classification models since it does not include a mapping of articles to a specific news taxonomy.

The main motivation for this work is to provide a dataset for building specific topic models. It consists of a categorized subset taken from an existing news dataset. We show that such a dataset, with up-to-date articles mapped into a standardized news taxonomy, can contribute to the accuracy improvement of news classification models.

\section{Methods}

In this paper, we present a new dataset based on the NELA-GT-2019 data source~\cite{Mauricio2019}, classified with IPTC's\endnote{The International Press Telecommunications Council, or IPTC, is an organization that creates and maintains standards for exchanging news and other information between news organizations.} NewsCodes Media Topic taxonomy~\cite{IPTCtaxonomy}. The original NELA-GT-2019 dataset contains 1.12 M news articles from 260 sources collected between  1 January 2019 and  31 December 2019, providing essential content diversity and topic coverage. Sources include a wide range of mainstream and alternative news outlets.

In turn, the IPTC taxonomies are a set of controlled vocabularies used to describe news stories’ content. The NewsCodes Media Topic taxonomy has been one of IPTC's main subject taxonomies for text classification since 2010. We used the 2020 version of NewsCodes Media Topic taxonomy~\cite{2020newscodes}. News organizations use it to categorize and index their content, while search engines use it to improve the discoverability of news stories~\cite{newcodesusage}.

Algorithm of the article selection process:
\begin{enumerate}
  \item Obtain a random article from the NELA dataset;
  \item Classify it for the second-level category of the NewsCodes Media Topic taxonomy by checking the keywords and thorough reading of the article; the news article is assigned to exactly one category;
  \item If there are already 100 articles in that category discard it, otherwise assign a second-level category to the article;
  \item Return to step 1 and repeat until each second-level category has 100 articles assigned.
\end{enumerate}

The described algorithm allows for overcoming the limitation of the NELA-GT datasets where a large proportion of the dataset is fringe, conspiracy-based news due to the discharging of the news if a category already has 100 articles in it.

We observed that the first-level category of the NewsCodes Media Topic taxonomy is not accurate enough to catalogue an article. For example, the ``sport'' category may include different aspects, such as information about specific sports, sports event announcements, and the sports industry in general, which have more specific meanings than the first-level category label is able to convey. Therefore, we used a second-level category of NewsCodes Media Topic taxonomy to have a more specific article category. In comparison to the previously published datasets, we included in our dataset unique categories such as ``arts and entertainment'', ``mass media'', ``armed conflict'', ``weather statistic'', and ``weather warning''. Therefore, we created the proposed Multilabeled News Dataset (MN-DS) by hand-picking and labeling approximately 100 news articles for each second level category\endnote{\url{https://www.iptc.org/std/NewsCodes/treeview/mediatopic/mediatopic-en-GB.html} (accessed on 13 March 2022)} of the NewsCodes Media Topic taxonomy.
\section{Data Records}

After manually selecting news articles relevant to each category, we obtained 10,917 articles in 17 first-level and 109 second-level categories from 215 media sources. During the selection process, one article was processed by one coder. An overview of the released MN-DS dataset by category is provided in Table~\ref{tab:tablenumberofarticles}. All data are available in CSV format at
\url{https://doi.org/10.5281/zenodo.7394850} under a Creative Commons license.
\vspace{-6pt}

\begin{table}[H]
\centering
    \caption{The number of articles under each Level 1~category.}
    \label{tab:tablenumberofarticles}
    \begin{tabularx}{\textwidth}{lC}
    \toprule
    \multicolumn{1}{c}{\textbf{Categories}} & {\textbf{Count}}\\
    \midrule
    Arts, culture, entertainment, and media  & 300\\
    Conflict, war, and peace & 800\\
    Crime, law, and justice & 500\\
    Disaster, accidents, and emergency incidents & 500\\
    Economy, business and finance  & 400\\
    Education  & 607\\
    Environment & 600\\
    Health  & 700\\
    Human interest & 600\\
    Labor  & 703\\
    Lifestyle and leisure & 300\\
    Politics & 900\\
    Religion and belief & 800\\
    Science and technology & 800\\
    Society  & 1100\\
    Sport & 907\\
    Weather & 400\\
    \bottomrule
    \end{tabularx}
\end{table}

The MN-DS contains articles published in 2019, the distribution of selected articles over the year is balanced with slightly more articles for the month of January 2019. The majority of the articles were selected from mainstream sources such as ABC News,
the BBC, The Sun, TASS, The Guardian, Birmingham Mail, The Independent, Evening Standard, and others. The dataset also includes a relatively small percentage of articles from alternative sources such as Sputnik, FREEDOMBUNKER, or Daily Buzz Live.

\begin{figure}[H]
        \centering
        \begin{subfigure}[b]{0.475\textwidth}
            \centering
            \includegraphics[width=\textwidth]{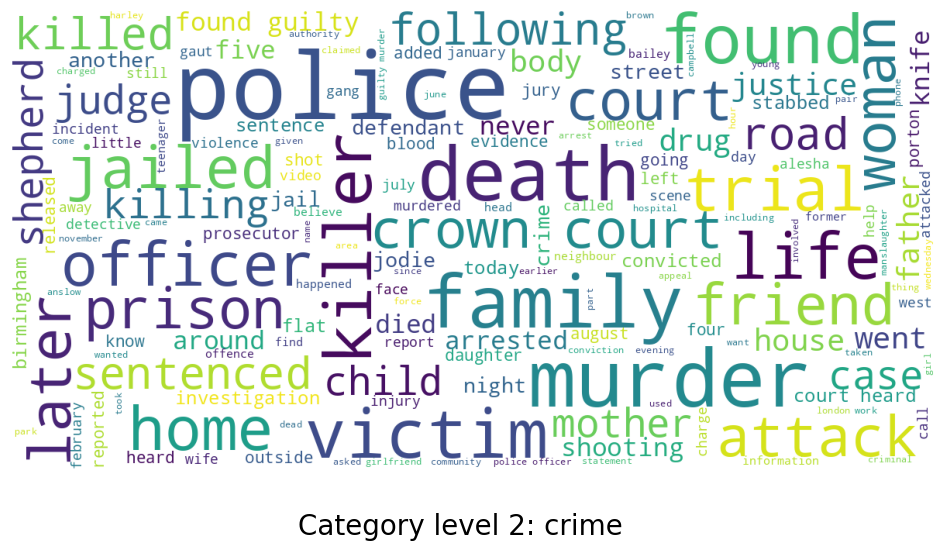}
        \end{subfigure}
        \hfill
        \begin{subfigure}[b]{0.475\textwidth}
            \centering
            \includegraphics[width=\textwidth]{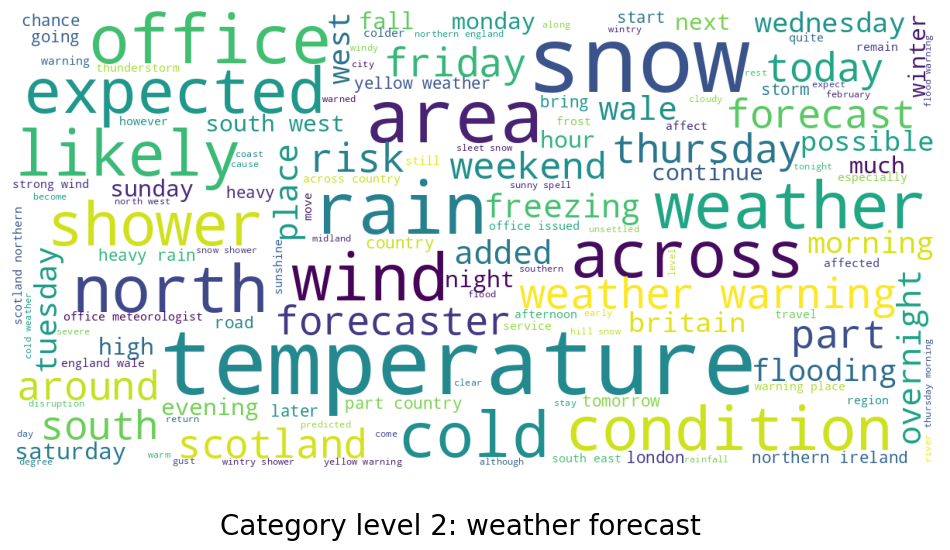}
        \end{subfigure}
        \vskip\baselineskip
        \begin{subfigure}[b]{0.475\textwidth}
            \centering
            \includegraphics[width=\textwidth]{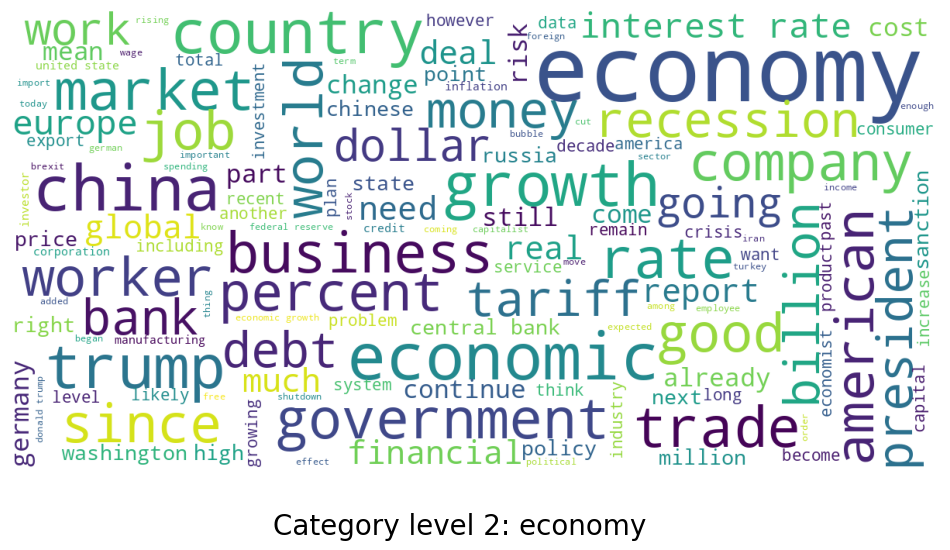}
        \end{subfigure}
        \hfill
        \begin{subfigure}[b]{0.475\textwidth}
            \centering
            \includegraphics[width=\textwidth]{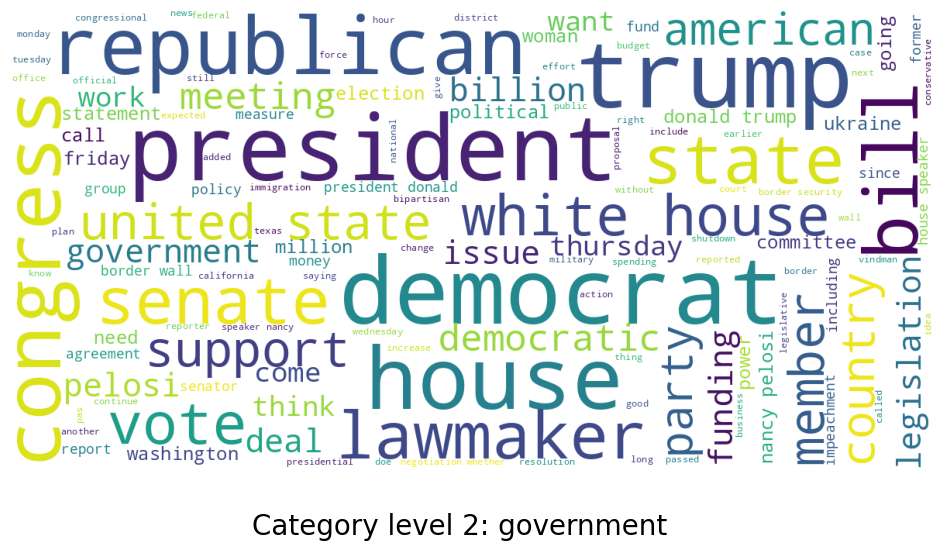}
        \end{subfigure}
        \caption{Word clouds of MN-DS dataset for selected second-level categories.}
        \label{figwordcloud}
\end{figure}

\begin{figure}[H]
\includegraphics[scale=0.4]{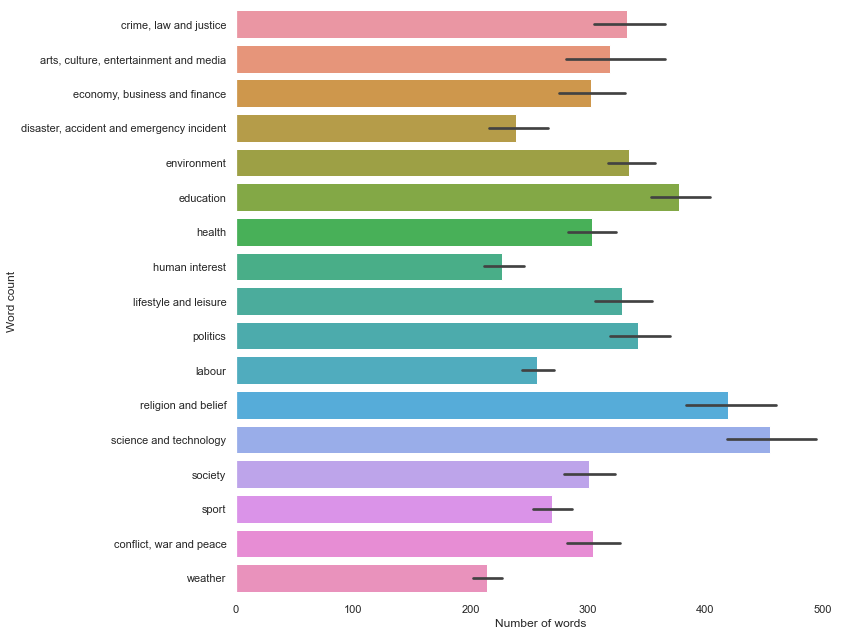}
\caption{Mean number of non-repeated words in article body for first-level categories. The error bars represent the 95\% confidence interval.}
\label{figmeannumbersofwordsnonrepeated}

\end{figure}

To describe the dataset, we created a word cloud representation of each category, as shown in Figure~\ref{figwordcloud}. The central concept of a word cloud is to visualize for each category the most popular words with a size corresponding to the degree of popularity. This representation allows us to quickly assess the quality of the text annotation since it displays the most common words of the category. In the bar chart shown in Figure~\ref{figmeannumbersofwordsnonrepeated}, we can observe that the ``science and technology'' first-level category contains the highest count of topic-specific words, while in more general categories, such as ``weather'' or ``human interest'', there is less variety in the texts, probably because they represent shorter and more similar articles.

The purpose of this dataset is to provide labeled data to train and test classifiers to predict the topic of a news article. Since the MN-DS represent the subset of the NELA-GT dataset, it could be also used to study the veracity of news articles but is not limited to this application. Due to the nature of the NELA-GT dataset, the style of articles is less formal, and we expect it to be the best fit for the alternative/conspiracy sources or social media article classification.

\subsection*{Description of Columns in the Data Table}

\begin{itemize}
\item id: Unique identifier of the article.
\item date: Date of the article release.
\item source: Publisher information of the article.
\item title: Title of the news article.
\item content: Text of the news article.
\item author: Author of the news article.
\item url: Link to the original article.
\item published: Date of article publication in local time.
\item published\_utc: Date of article publication in utc time.
\item collection\_utc: Date of article scraping in utc time.
\item category\_level\_1: First level category of Media Topic NewsCodes's taxonomy.
\item category\_level\_2: Second level category of Media Topic NewsCodes's taxonomy.
\end{itemize}
\vspace{-9pt}

\section{Usage Example}

We used the dataset to train the most common text classification models to extend the technical validation of the proposed dataset and establish the benchmark for multiclass classification. The following embeddings were selected:

\begin{itemize}
\item Tf-idf embedding, where Tf-idf stands for term frequency-inverse document frequency~\cite{linktfidf}. Tf-idf transforms text into a numerical representation called a tf-idf matrix. The term frequency is the number of times a word appears in a document. The inverse document frequency measures how common a word is across all documents. Tf-idf is used to weigh words so that important words are given more weight. The dataset's news texts and categories were combined and vectorized with TfidfVectorizer~\cite{sklearnref}.
\item GloVe (Global Vectors for Word Representation) embeddings with an algorithm based on a co-occurrence matrix, which counts how often words appear together in a text corpus. The resulting vectors are then transformed into a lower-dimensional space using singular value decomposition~\cite{linkglove}.
\item DistilBertTokenizer~\cite{linkbert}, which is a distilled version of BERT, a popular pre-trained model for natural language processing. DistilBERT is smaller and faster than BERT, making it more suitable for fast training with limited resources. The trade-off is that DistilBERT's performance is 3\% lower than BERT's. DistilBERT embeddings are trained on the same data as BERT, so they are equally good at capturing the meaning of words in context.

\end{itemize}

During dataset validation, we combined the selected embeddings with different classifiers. We tested multinomial naive Bayes (NB) classifier~\cite{linkmnb}, logistic regression~\cite{linklogisticregression}, support vector classifier (SVC)~\cite{linksvm}, and DistilBERT model~\cite{distilbert}. Since MN-DS is a multiclass dataset, we used the OneVsRestClassifier strategy for classification models~\cite{sklearnref}. OneVsRestClassifier is a classifier that trains multiple binary classifiers, one for each class. The individual binary classifiers are then combined to create a single multiclass classifier. This approach is often used when there are many categories, as it can be more efficient than training a single multiclass classifier from scratch. The tested classifiers work as follows:

\begin{itemize}
\item The multinomial NB is a text classification algorithm that uses Bayesian inference to classify text. It is a simple and effective technique that can be used for various tasks, such as spam filtering and document classification. The algorithm is based on the assumption that the features in a document are independent of each other, which allows it to make predictions about the category of a document based on its individual~features.

\item The logistic regression classifier works by using a sigmoid function to map data points from an input space to an output space, where the categories are assigned based on a linear combination of the features. The weights of the features are learned through training, and the predictions are made by taking the dot product of the feature vector and the weight vector.

\item The SVC classifier is a powerful machine learning model based on the support vector machines algorithm. The model is based on finding the optimal decision boundary between categories to maximize the margin of separation between them. The SVC model can be used for linear and non-linear classification tasks and is particularly well-suited for problems with high dimensional data. The classifier is also robust to overfitting and can generalize well to new data.

\item DistilBERTModel, a light version of the BERT classifier~\cite{linkbert}, developed and open-sourced by the team at Hugging Face. DistilBERTModel can be fine-tuned with just one additional output layer to create state-of-the-art models for a wide range of NLP tasks with minimal training data.
\end{itemize}

Classification results for level 1 and level 2 categories are presented in Table~\ref{tab:tableresults} and Table~\ref{tab:tableresultssecond}, respectively. It is possible to observe that DistilBERTModel achieves better classification results for both category levels. To improve these results in future studies, we suggest applying hierarchical classification methods as described by Silla and  Freitas~\cite{linkhierarhical}, for example.

\vspace{-3pt}

\begin{table}[H]
\centering
    \caption{Multilabel classification results for level 1~categories.}
    \label{tab:tableresults}

\begin{adjustwidth}{-\extralength}{0cm}
\begin{tabularx}{\fulllength}{cCCCCCCCCC}
    \toprule
    \multicolumn{1}{c}{\textbf{Embeddings}} & \multicolumn{3}{c}{\textbf{TFIDF}} & \multicolumn{3}{c}{\textbf{Glove}} & \multicolumn{3}{c}{\textbf{DistilBertTokenizer}}\\
    \cmidrule(r){2-4}\cmidrule(r){5-7}\cmidrule(r){8-10}
    \multicolumn{1}{c}{\textbf{Model}} & \textbf{Precision} & \textbf{Recall} & \textbf{f1 Score} & \textbf{Precision} & \textbf{Recall} &\textbf{ f1 Score} & \textbf{Precision} & \textbf{Recall} & \textbf{f1 Score} \\
    \midrule
    Multinomial NB  & 0.802  & 0.631 & 0.649 & 0.629 & 0.499 & 0.529 & n/a & n/a & n/a \\
    Logistic Regression& 0.800  & 0.763 & 0.774 & 0.747 & 0.739 & 0.739 & n/a & n/a & n/a \\
    SVC Classifier    & 0.808  & 0.796 & 0.799 & 0.768 & 0.762 & 0.760 & n/a & n/a & n/a \\
    DistilBERTModel   & n/a & n/a & n/a & n/a & n/a & n/a & 0.849 & 0.842 & 0.844 \\
    \bottomrule
    \end{tabularx}
\end{adjustwidth}
\end{table}
\unskip

\begin{table}[H]
\centering
    \caption{Multilabel classification results for level 2~categories.}
    \label{tab:tableresultssecond}

\begin{adjustwidth}{-\extralength}{0cm}
   \begin{tabularx}{\fulllength}{cCCCCCCCCC}
    \toprule
    \multicolumn{1}{c}{\textbf{Embeddings}} & \multicolumn{3}{c}{\textbf{TFIDF}} & \multicolumn{3}{c}{\textbf{Glove}} & \multicolumn{3}{c}{\textbf{DistilBertTokenizer}}\\
    \cmidrule(r){2-4}\cmidrule(r){5-7}\cmidrule(r){8-10}
    \multicolumn{1}{c}{\textbf{Model}} & \textbf{Precision} & \textbf{Recall} & \textbf{f1 Score} & \textbf{Precision} & \textbf{Recall} & \textbf{f1 Score} & \textbf{Precision} & \textbf{Recall} & \textbf{f1 Score} \\
    \midrule
    Multinomial NB  & 0.628  & 0.602 & 0.583 & 0.496 & 0.484 & 0.469 & n/a & n/a & n/a \\
    Logistic Regression & 0.646  & 0.649 & 0.635 & 0.589 & 0.589 & 0.577 & n/a & n/a & n/a \\
    SVC Classifier    & 0.645  & 0.646 & 0.628 & 0.581 & 0.595 & 0.571 & n/a & n/a & n/a \\
    DistilBERTModel   & n/a & n/a & n/a & n/a & n/a & n/a & 0.735 & 0.715 & 0.715 \\
    \bottomrule
    \end{tabularx}
\end{adjustwidth}
\end{table}

\vspace{-6pt}

\authorcontributions{A.P. created the search strategy, retrieved and screened the publications, extracted the selected data for annotation, and labeled the dataset. A.P. and N.F. assessed the quality of the included articles and checked the data. A.P. performed the statistical analyses, created the graphics and wrote the original draft. N.F. conceived the project, provided critical comments, and revised the paper. All authors have read and agreed to the published version of the manuscript.}

\funding{This research was funded by Fundação para a Ciência e Tecnologia under Project UIDB/04111/2020 (COPELABS).}

 \institutionalreview{
 Not applicable
}

 \informedconsent{
 Not applicable
}

\dataavailability{The data described in this paper is available in CSV format at \url{https://\\doi.org/10.5281/zenodo.7394850}.
Code for the technical validation of the dataset is available at \url{https://github.com/alinapetukhova/mn-ds-news-classification} (accessed on 15 April 2023).}

\conflictsofinterest{The authors declare no conflicts of interest.}

\begin{adjustwidth}{-\extralength}{0cm}

\printendnotes[custom]

\reftitle{References}

\PublishersNote{}
\end{adjustwidth}
\end{document}